\documentclass[journal]{IEEEtran}

\usepackage{amsmath,amssymb}
\usepackage{balance}
\usepackage{color}
\usepackage{cite}
\usepackage{graphicx}
\usepackage{multirow}
\usepackage{url}
\usepackage{caption}

\ifCLASSINFOpdf
\else
\fi

\begin{document}

\title{Crowd Counting with Density Adaption Networks}

\author{Li Wang, Weiyuan Shao, Yao Lu, Hao Ye, Jian Pu, Yingbin Zheng
\vspace{-0.3in}
\thanks{L. Wang, W. Shao, H. Ye, and Y. Zheng are with Shanghai Advanced Research Institute, Chinese Academy of Sciences, Shanghai, China. Y. Lu is with University of Washington, Seattle, WA, USA. J. Pu is with East China Normal University, Shanghai, China.
}
}

\maketitle

\begin{abstract}

Crowd counting is one of the core tasks in various surveillance applications. A practical system involves estimating accurate head counts in dynamic scenarios under different lightning, camera perspective and occlusion states. Previous approaches estimate head counts despite that they can vary dramatically in different density settings; the crowd is often unevenly distributed and the results are therefore unsatisfactory. In this paper, we propose a lightweight deep learning framework that can automatically estimate the crowd density level and adaptively choose between different counter networks that are explicitly trained for different density domains. Experiments on two recent crowd counting datasets, UCF\_CC\_50 and ShanghaiTech, show that the proposed mechanism achieves promising improvements over state-of-the-art methods. Moreover, runtime speed is 20 FPS on a single GPU. 

\end{abstract}


\IEEEpeerreviewmaketitle

\section{Introduction}
With the growing deployment of surveillance video cameras, the surveillance applications have received increasing attention from the research community.
Different from other problems such as face and vehicle detection \cite{2016spl_kzhang,wang2017evolving,lyu2017ua},
crowd counting, or estimating the head count, from video frames, has been proved to be a critical functionality in various traffic and public security scenarios \cite{Zhang:2015id,Zhang:2016fr,sam2017switching,Sindagi:2017vv,zhang2017fcn,Xiong:2017ug,kumagai2017mixture}, and applications in zoology \cite{arteta2016counting} and neural science \cite{Xie:2015wr} further extend the usability and importance of the problem.

\begin{figure}[t]
  \centering
  \includegraphics[width=.7\linewidth]{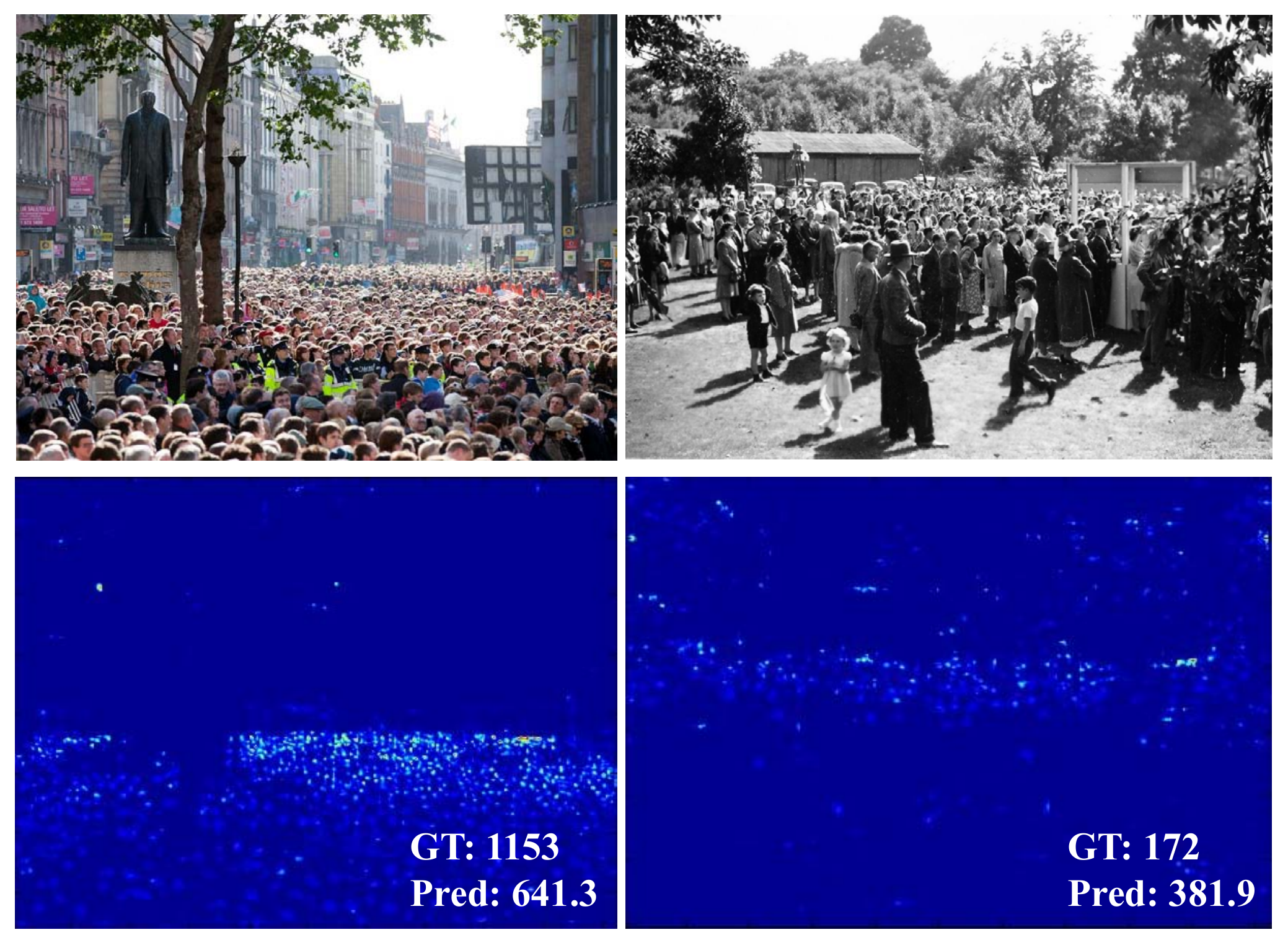}
  \caption{Examples of diverse crowds. First row: input images from ShanghaiTech dataset Part A. Second row: estimated density map. The unevenly-distributed crowd and the cluttered scenarios are the major challenges for accurate crowd counting. GT: ground truth. Pred: estimated count.}
  \label{fig:introductiongt}
  \vspace{-0.15in}
\end{figure}

Multiple challenges exist to produce accurate and efficient crowd counting results.
Heavy occlusion, lightning and camera perspective changes are the common issues. Moreover, as shown in Figure \ref{fig:introductiongt}, head counts vary dramatically in different scenarios; area of the head ranges from hundreds of pixels to only a few pixels, and the crowd is often unevenly distributed due to camera perspective or physical barriers. Recent approaches provide rough head count estimates based on multi-scale and context-aware cues \cite{Zhang:2015id,Zhang:2016fr,kumagai2017mixture,sam2017switching,Sindagi:2017vv,zhang2017fcn,Xiong:2017ug}; a universal normalization or scaling mechanism is often used for different density domains. Practically, this can be suboptimal and leads to inaccuracy especially when the scenario is highly dynamic. Better adaption to different density domains is a first-order question for current crowd counting algorithms.

In this paper, we propose a deep learning pipeline to automatically infer the crowd density level given a single input image, and our framework adaptively chooses a counter network that is explicitly trained for the target density domain. Different counter networks and the density level estimator are associated in a spatial gating unit for end-to-end crowd counting. With this ground, our framework addresses the density adaption problem and produces more satisfactory results.
The idea of constructing representation from multiple levels has been taken in several previous works, e.g., the Switch-CNN \cite{sam2017switching}. Compared to these approaches using different network structures as regressors and classifier, our subnetworks employ the similar design and are easy to train.

To evaluate our proposed framework, we report the evaluations
on the recent ShanghaiTech \cite{Zhang:2016fr} and the UCF\_CC\_50 crowd counting dataset \cite{idrees2013multi}, and we compare with several recent approaches including MCNN \cite{Zhang:2016fr} and CP-CNN \cite{Sindagi:2017vv}. Notably, we
achieve a significant 35.8 MAE improvements over the state-of-the-art Shang et al. \cite{Shang:2016iq} on UCF\_CC\_50, and 2.5 MAE gains over CP-CNN \cite{Sindagi:2017vv} on ShanghaiTech Part B. Meanwhile, a 20 FPS processing speed is obtained on an Nvidia Titan X GPU (Maxwell).

\subsection{Related Work}
\label{sec:rw}
A number of studies on crowd counting have been demonstrated to solve the real world problem \cite{loy2013crowd,sindagi2017survey}. They can be summarized into three categories depending on the methodology: detection-based, regression-based and density-based, which will be briefly reviewed below.

\vspace{0.05in}\noindent \textbf{Detection-based crowd counting} is straightforward and utilizes off-the-shelf detectors \cite{li2008estimating,leibe2005pedestrian,wang2009crowd} to detect and count target objects in images or videos. However, for crowded scenarios, objects are highly occluded and many objects are too small to detect. All these make the counting inaccurate.

\vspace{0.05in}\noindent \textbf{Regression-based crowd counting}. Regression-based approaches such as \cite{chan2009bayesian,chen2012feature,Segui:2015ho,CarlosArteta:2014vm,Shang:2016iq, kumagai2017mixture,wang2017deep} are proposed to bypass the occlusion problem that can be critical for detection-based methods. Specifically, a mapping between image features and the head count is recovered, and the system benefits from better feature extraction and count number regression algorithms \cite{chen2012feature, Segui:2015ho,Shang:2016iq,chan2009bayesian, CarlosArteta:2014vm}. Moreover, \cite{chan2008privacy, ryan2009crowd,chan2012counting,fu2012real} leverage spatial or depth information and use segmentation methods to filter the background region and regress count numbers only on foreground segments. These type of methods are sensitive to different crowd density levels and heavily depend on a normalization strategy that is universally good.

\vspace{0.05in}\noindent \textbf{Density-based crowd counting.} \cite{lempitsky2010learning,Zhang:2015id,Zhang:2016fr,sam2017switching,Sindagi:2017vv,zhang2017fcn} use continuous density maps instead of discrete count numbers as the optimization objective and learn a mapping between the image feature and the density map. Specifically, \cite{Zhang:2015id} presents a data-driven method to count in unseen scenarios. \cite{Zhang:2016fr} proposes a multi-column network to directly generate the density map from input images. \cite{walach2016learning} introduces the boosting process which yield a significant improvement both in accuracy and runtime. To address perspective-free problem, \cite{onoro2016towards} feeds a pyramid of input patches into their own designed network. \cite{sam2017switching} improves over \cite{Zhang:2016fr} and uses a switch layer to exploit the variation of the crowd density. \cite{zhang2017fcn} jointly estimates the density map and count number with FCN and LSTM layers. \cite{Sindagi:2017vv} uses global and local context to generate high quality density map. The insufficiency of these type of methods is that the mapping between density and image may lead to deviation and the actual count can often be inaccurate.

In this paper, our framework leverages both continuous density map and discrete head count annotations in training; a density-level domain adaption network is used to explicitly recognize the domain allocation of each image patch. Different from previous works, we do not focus on pursuing a better density estimation but count directly. It totally drops the local details and is hard to learn. Instead, we propose a count map, which to some extent preserves the local details and can be calculated analytically. Besides, if we only use a single network to predict count map, the result will be dominated by the patch with high-density crowds. Therefore, we classify each patch into low- or high-density. This step has two advantages: i) counting number of patches with low-density crowds becomes more accurate; ii) the classifier becomes easier to train.

\section{Framework}
\label{sec:method}

\begin{figure}[t]
  \centering
  \includegraphics[width=.99\linewidth]{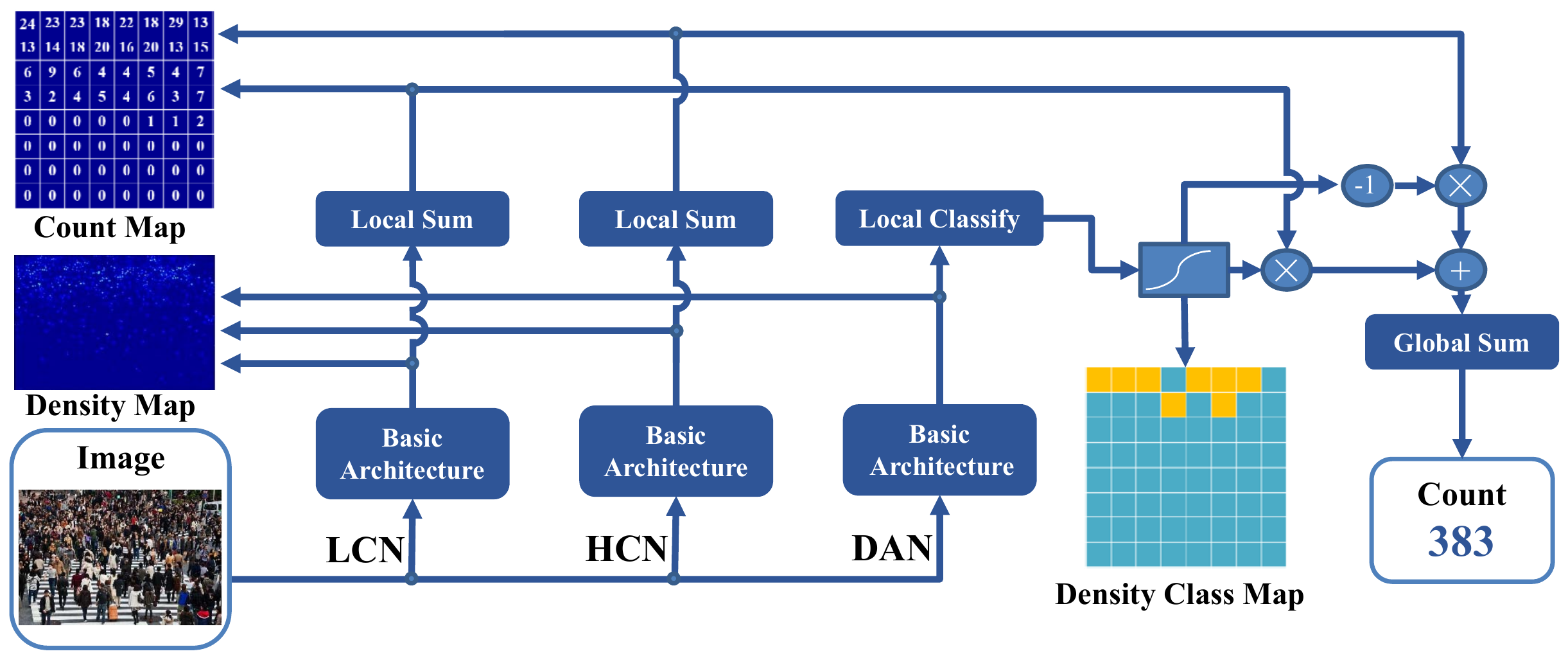}
  \caption{The overall architecture of our framework. Our proposed framework contains a Density Adaption Network (DAN), a Low-density Counter Network (LCN) and a High-density Counter Network (HCN). The three networks share the similar base structure, DAN generate class map for an image to identify each patch with low- or high-density while LCN and HCN are used to generate accurate head count maps. And the final results is the integration of count maps switching between LCN and HCN.}
  \label{fig:architecture}
  \vspace{-0.15in}
\end{figure}

The overall architecture of our framework is demonstrated in Figure \ref{fig:architecture}. Due to the dynamic crowd density between images and patches, we propose an adaptive network structure that includes a Density Adaption Network (DAN), a Low-density Counter Network (LCN) and a High-density Counter Network (HCN). DAN identifies each patch in the image if they belong to the low- or high-density domain; LCN and HCN are used to generate accurate head counts for low- or high-density patches. We note that using only the continuous density map or the discrete count map often leads to inaccurate estimation results. Instead, both of them will be considered in our framework.

\begin{table}[t]
\begin{center}
  \caption{The architecture of proposed network. }
  \label{table:network}
  \begin{tabular}{cccc}
    \hline
    \multirow{2}{*}{Type} & Kernel Size / & In/Output & \multirow{2}{*}{\#Params}\\
    & Stride / Dilation & Channels & \\\hline
    {\tt conv1\_1} & $3\times 3$ / 1 / 1 & $(3,24) $ & 2.6$K$\\
    {\tt conv1\_2} & $3\times 3$ / 1 / 1 & $(24,24)$ & 20.3$K$\\
    {\tt conv1\_3} & $3\times 3$ / 1 / 1 & $(24,24)$ & 20.3$K$\\
    {\tt conv2} & $3\times 3$ / 1 / 2 & $(24,48)$ & 40.7$K$\\
    {\tt conv3} & $3\times 3$ / 1 / 4 &  $(48,24)$  & 40.6$K$\\
    {\tt conv4} & $3\times 3$ / 1 / 2 &  $(24,12)$ & 10.2$K$\\
    {\tt conv5} & $3\times 3$ / 1 / 1 &  $(12,12)$  & 5.1$K$\\
    \hline
    {\tt density map} & $1\times 1$ / 1 / 1 & $(12,1)$ & 0.05$K$\\
    \hline
    \multirow{2}*{LCN/HCN} & {\tt density map}&&\\
    & $64\times 64$ / 64 / 1 & $(1,1)$ & 16.0$K$\\

    \hline
   \multirow{2}*{DAN} & $64\times 64$ / 64 / 1 & $(12,12)$ & 2304$K$\\
     & $1\times 1$ / 1 / 1 & $(12,2)$ & 0.1$K$\\
    \hline
  \end{tabular}
  \end{center}
  Here we suppose the input image size is $512\times 512$. Please note that {\tt conv1-5} correspond to the \textbf{Basic Architecture} in Figure \ref{fig:architecture}, the layers in LCN/HCN and DAN correspond to the \textbf{Local Sum} and \textbf{Local Classify} in Figure \ref{fig:architecture} respectively.
  \vspace{-0.15in}
\end{table}

As shown in Figure \ref{fig:architecture} and Table \ref{table:network}, each network uses a similar CNN architecture to estimate the density map $\hat D = \sum_{c} \mathcal{N}(x_{c},y_{c},\sigma_{x_{c}}^2,\sigma_{y_{c}}^2)$ with $c$ independent Gaussian centered at each head. \footnote{The symbols with hat denote the estimation of networks, while symbols without hat represent the ground truth.} $\sigma_i^2$ is pre-defined as in \cite{Zhang:2016fr}. Like \cite{idrees2013multi}, we divide the input image into $H\times W$ grids. LCN leverages one additional convolutional layer to generate the count map $\hat C_{lcn}^{i,j} = \sum_{(x,y)\in grid(i,j)} D(x,y)$ locally, and $\hat C_{hcn}^{i,j}$ and $\hat C_{dan}^{i,j}$ are defined similarly for HCN and DAN. We polarize the DAN output and generate the density class by:

\begin{equation}
\hat P^{i,j}=\left\{
\begin{aligned}
0,&& C_{dan}^{i,j}\leq th \\
1,&& \text{otherwise}
\end{aligned}
\right.
\end{equation}

\noindent where $th$ is a threshold, and therefore the spatial gating unit switches between LCN and HCN and the final head count in the image is given by:
\begin{equation}
\hat C = \sum_{i,j} (\hat C_{lcn}^{i,j} \otimes (1 - \hat P^{i,j}) + \hat C_{hcn}^{i,j}\otimes \hat P^{i,j})
\end{equation}

\noindent where $\otimes$ is element-wise production.

\vspace{0.05in}\noindent \textbf{Training strategy}.
We note that although the structures for DAN, LCN and HCN are similar, they are recognized as different parts of the network. This is because they are explicitly trained from different density domains; multiple supervision signals are involved in the training. We observe in our experiments that DAN as well as the two counters will produce reasonable results only with good initiations. Therefore, we start training the Basic Architecture only with density map annotations; after convergence, density class $\hat P$ and count map $\hat C$ annotations are added to fine-tune DAN, LCN and HCN on previous basic model. To convert a density map into a count map in our network inference, we use a consecutive convolutional layer, which we expect after a few epochs can learn a better mapping compared with a simple sum pooling. Losses for the density map, head count and density class for each network unit are defined as follows:
\begin{equation}
  L_{density}(\theta) = \frac{1}{2N} \sum_{i=1}^{N} || \hat D(X_i;\theta) - D_i ||^2\label{densloss},
\end{equation}
\vspace{-0.05in}
\begin{equation}
  L_{count}(\theta) = \frac{1}{N} \sum_{i=1}^{N} ||\hat C(X_i;\theta) - C_i ||\label{countloss},
\end{equation}
\vspace{-0.05in}
\begin{equation}
  L_{p}(\theta) = - \frac{1}{N}[P_i\log(\hat P(X_i)) + (1 - P_i)\log(1 - \hat P(X_i))]\label{clsloss}
\end{equation}

\noindent where $\theta$ is the network parameters, $N$ is the number of training samples and $X_i$ is the $i$th image. $D_i$, $C_i$, and $P_i$ are the ground truth density map, count map and density class respectively\footnote{It's worth noting that Least square errors (L2) and Least absolute deviations (L1) are applied to Eq. (\ref{densloss}), (\ref{countloss}) respectively. $|\hat C(X_i;\theta) - C_i |$ is usually much larger than $| \hat D(X_i;\theta) - D_i |$, meaning that head count loss is more sensitive to outlier; and the variance will be further augmented and the sample with thousands of people may dominate the final function if L2 loss is employed in head count loss function.}. Therefore, the multi-part loss functions for different network parts are defined as combination of the density and count (density class) losses:
\begin{equation}
  L^{dan}(\theta) = L_{density}^{dan}(\theta) + \lambda_{dan}L_{dan}(\theta), \label{cnetloss}
\end{equation}
\begin{equation}
  L^{lcn}(\theta) = L_{density}^{lcn}(\theta) + \lambda_{lcn}L_{count}^{lcn}(\theta), \label{snetloss}
\end{equation}
\begin{equation}
  L^{hcn}(\theta) = L_{density}^{hcn}(\theta) + \lambda_{hcn}L_{count}^{hcn}(\theta).  \label{dnetloss}
\end{equation}

\vspace{0.05in}\noindent \textbf{Implementation details}. We implement our framework in PyTorch\footnote{\url{http://pytorch.org/}} and we note a few practical issues here. First, all input images are resized to $512\times512$ with 3 channels as the network input and the aspect ratio are kept with zero padding. The first three layers in {\tt conv1} are three consecutive $3\times3$ convolutional layers. As dilated convolutional layers have been shown to be effective in many computer vision tasks \cite{yu2017dilated,zheng2018learning,rene2017temporal,xu2018dense}, the following 4 layers ({\tt conv 2-5}) are dilated layers with dilation parameter = 2,4,2,1 respectively. The last $1\times 1$ convolutional layer is a bottleneck to regress the density level. Second, in order to choose the appropriate thresholds for different datasets, we add up the count values of patches in training images and the final thresholds are set according to the intermediate value of the statistics. Then, DAN connects two consecutive convolutional layers after {\tt conv5}, the output serves as the density gate with size $8 \times 8$. LCN and HCN connect one consecutive convolutional layers after {\tt density map} layer to obtain the corresponding count maps. Finally, we augment the training data with only random flips and we use Adam with learning rate=$10^{-5}$.

\section{Experiments}
\label{sec:exp}

We demonstrate crowd counting results compared with previous works on two recent datasets: the ShanghaiTech Dataset \cite{Zhang:2016fr} and the UCF\_CC\_50 Dataset \cite{idrees2013multi}. The effectiveness of each component in our module is evaluated and the influence of density class map size will be explored. We further consider transferring the learning between the datasets.

The metric we use include Mean Absolute Error, MAE=$\frac{1}{N} \sum_{i=1}^{N} | C_i - \widehat C_i |$, and Mean Squared Error, MSE=$\sqrt{\frac{1}{N}  \sum_{i=1}^{N} ( C_i - \widehat C_i )^2}$, where $N$ is the number of the testing images and $C_i$ and $\widehat C_i$ are the ground truth and the predicted count number in the $i$-th test image.

\subsection{Datasets and Results}

\noindent \textbf{The UCF\_CC\_50 Dataset} \cite{idrees2013multi} contains 50 images with head counts ranging from 94 to 4,543, and a total number of 63,974 individuals are annotated. Despite that the number of images is not large, the diversity of the scenarios makes the dataset extremely challenging. We conduct a five-fold cross-validation for training and testing, which is the standard evaluation setting used in \cite{idrees2013multi}. In the training, we generate the density map using the same spread ($\sigma$) in the Gaussian kernel, and the threshold $th$ for the density boundary that decides the patch sparcity is set to 40 due to the high-density crowd in the dataset.

\begin{figure*}[t]
  \centering
  \includegraphics[width=.75\linewidth]{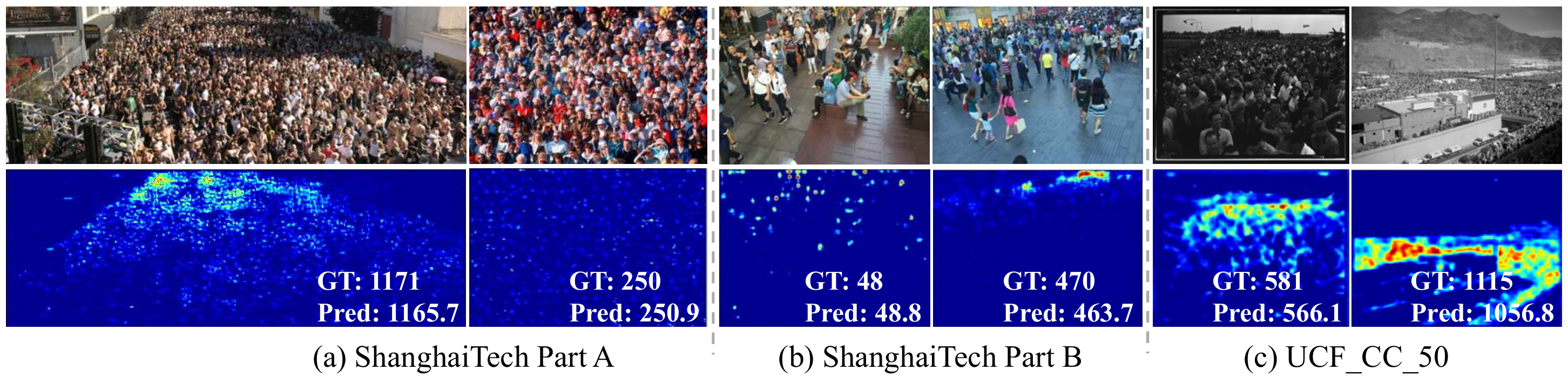}
  \caption{Qualitative results on the benchmarks.}
  \label{fig:further_results}
\vspace{-0.15in}
\end{figure*}

Table \ref{tab:ucf} shows the results on UCF\_CC\_50. We compare with \cite{Zhang:2015id, Zhang:2016fr, sam2017switching, Sindagi:2017vv, Shang:2016iq} that are state-of-the-art CNN-based approaches, except for \cite{Shang:2016iq} that uses LSTM over a sequence of video frames. Recall that Shang et al. \cite{Shang:2016iq} use additional training images, and our method still achieves state-of-the-art MAE and MSE, as our networks can leverage different density level patches to their appropriate counter and achieve the more accurate results. Examples of the testing results can be seen in Figure \ref{fig:further_results}(c).

\begin{table}
\begin{center}
\caption{Comparison on the UCF\_CC\_50 dataset.}
\label{tab:ucf}
\begin{tabular}{c|cc }
\hline
Method &MAE&MSE\\
  \hline
  \hline
  Zhang et al. \cite{Zhang:2015id} &	 467.0  &  498.5    \\
  \hline
  MCNN \cite{Zhang:2016fr} &	377.6   &  509.1   \\
  \hline
  Switching-CNN \cite{sam2017switching} &	318.1   & 439.2    \\
  \hline

  CP-CNN \cite{Sindagi:2017vv} &	295.8 &  320.9   \\
  \hline
  ConvLSTM \cite{Xiong:2017ug}  &	284.5   &  297.1  \\
  \hline
  Shang et al. \cite{Shang:2016iq}  &	270.3   &   -  \\
  \hline
  Our Method   &	\textbf{234.5}   &  \textbf{289.6}   \\
  \hline
\end{tabular}
\end{center}
\vspace{-0.15in}
\end{table}

\vspace{0.05in}\noindent \textbf{The ShanghaiTech Dataset} \cite{Zhang:2016fr} is one of the largest datasets available in terms of annotation. It contains 1,198 annotated images with a total of 330,165 people. The dataset consists of two subsets: Part A and Part B. Part A has 482 images collected from the Internet while Part B includes 716 images captured from downtown Shanghai. The dynamic scenarios make the dataset even more challenging. We conduct our experiments following setting of \cite{Zhang:2016fr} where Part A is divided into 300 images for training and 182 images for testing, and 400 images in Part B for training and the rest for testing. In the training process, we generate the ground truth density map as in \cite{Zhang:2016fr} with geometry-adaptive kernels for Part A and the same spread in Gaussian kernel for Part B. The density boundary threshold $th$ is set to 20 for Part A and 10 for Part B.

\begin{table}
\begin{center}
\caption{Comparison on the ShanghaiTech dataset.}
\label{tab:shanghaitech}
\begin{tabular}{c|cc|cc }
\hline
\multirow{2}{*}{Method} &\multicolumn{2}{c|}{Part A}&\multicolumn{2}{c}{Part B}\\
\cline{2-5}
&MAE&MSE&MAE&MSE\\
  \hline
  \hline
  Zhang et al. \cite{Zhang:2015id} &	 181.8  &  277.7  &  32.0  &  49.8  \\
  \hline
  MCNN \cite{Zhang:2016fr} &	110.2   &  173.2  & 26.4   & 41.3   \\
  \hline
  Switching-CNN \cite{sam2017switching} &	90.4   & 135.0   &  21.6  &  33.4  \\
  \hline
  CP-CNN \cite{Sindagi:2017vv} &	\textbf{73.6}  &  \textbf{106.4}  &  20.1  &  30.1  \\
  \hline
  Our Method  &	 88.5  &  147.6  &  \textbf{17.6}  &  \textbf{26.8}  \\
  \hline
\end{tabular}
\end{center}
\vspace{-0.15in}
\end{table}

Table \ref{tab:shanghaitech} demonstrates the comparison of our model with state-of-the-art approaches on the ShanghaiTech dataset: Zhang et al. \cite{Zhang:2015id}, MCNN \cite{Zhang:2016fr}, Switching-CNN \cite{sam2017switching} and CP-CNN \cite{Sindagi:2017vv}. Our approach achieves a promising improvement of 2.5 MAE and 3.3 MSE on Part B while producing comparable results on Part A. We note that our network structure is much simpler than the CP-CNN and hence is much faster.
Our framework runs at 20FPS on an Nvidia X GPU (Maxwell), and qualitative results on Part A and Part B can be seen in Figure \ref{fig:further_results}(a) and \ref{fig:further_results}(b).

\subsection{Ablation Study}

\begin{table}
\begin{center}
\caption{Ablation experiments results on the exploring of three stream networks on UCF\_CC\_50.}
\label{tab:networks}
\begin{tabular}{c|cc}
\hline
Method &MAE&MSE\\
  \hline
  \hline
    LCN   &	660.9   &  867.6    \\
  \hline
  HCN   &	647.2  &  747.5   \\
  \hline
  LCN + HCN + DAN  &	\textbf{234.5}   &  \textbf{289.6} \\
  \hline\hline
  Ideal classification & 157.3  &  195.6 \\
  \hline
\end{tabular}
\end{center}
\vspace{-0.15in}
\end{table}

\begin{table}[t]
\begin{center}
\caption{MAE/MSE based on different grid sizes and input sizes on ShanghaiTech Part A.}
\label{tab:gridsize}
\begin{tabular}{c|cc|cc|cc}
\hline
Grid size &\multicolumn{2}{|c|}{4 $\times$ 4}&\multicolumn{2}{|c|}{8 $\times$ 8}&\multicolumn{2}{|c}{16 $\times$ 16}\\
\hline\hline
Input size &MAE&MSE&MAE&MSE&MAE&MSE\\
  \hline
    256 $\times$ 256  &	97.1 & 165.5   &  92.2 & 150.6&105.3 & 181.2   \\
  \hline
  512 $\times$ 512   &	100.0 & 180.5 &  88.5 & 147.6&99.8 & 155.3   \\
  \hline
 \end{tabular}
\end{center}
\vspace{-0.15in}
\end{table}

The effectiveness of each part will be evaluated in this section.
As shown in Table \ref{tab:networks}, LCN and HCN demonstrate MAE/MSE on the UCF\_CC\_50 dataset in which we use only LCN or HCN for crowd counting; using the counter from a single density domain produces much worse counting results due to the lack of context from other density level.
The DAN in our framework achieves the density domain allocation accuracy of 0.96.
The LCN + HCN + DAN demonstrates the performance combining LCN and HCN according to the classification results of DAN, while the last row shows the MAE/MSE using the ground truth density level rather than the predicted class map from DAN. It is clear that the density domain is a critical factor and there is still a gap between our results and optimal.

We also demonstrate the how the density class map grid size affects the results. We can see from the Table \ref{tab:gridsize} that the best performance can be get when  grid size and input size are set to 8$\times$8 and 512$\times$512 respectively.
Note that the grid size of 16$\times$16 is suboptimal for images with few heads due to division of a head into several parts. The grid size of 16$\times$16 is suboptimal for images with high-density crowd due to loss of local details. Overall, input size of 512$\times$512 preserves more details and achieves better results than that of 256$\times$256. We have also tried larger input sizes but the training becomes suboptimal.

\begin{table}[t]
\begin{center}
\caption{Transfer learning with different training strategies. We use ShanghaiTech Part A as the source domain and report the testing results on UCF\_CC\_50.}
\label{tab:transfer}
\begin{tabular}{c|cc }
\hline
Method &MAE&MSE\\
  \hline
  \hline
  W/O fine-tune      &	378.2   &  434.9 \\
  \hline

  Step learning on target & 234.5   &  289.6 \\

  \hline
  Fine-tune on target & \textbf{228.9}   &  \textbf{283.2}   \\
  \hline
  MCNN \cite{Zhang:2016fr} & 295.1  & 490.2   \\
  \hline
\end{tabular}
\end{center}
\vspace{-0.2in}
\end{table}

\subsection{Dataset Transfer}
We wonder how generalizable our proposed framework is. Similar to Zhang et al. \cite{Zhang:2016fr}, we verify the dataset transfer by using ShanghaiTech Part A as the source domain and UCF\_CC\_50 as the target domain.  The results are reported in Table \ref{tab:transfer}. We compare three training strategies. (i) W/O fine-tune: we use the base model pre-trained on the source domain only with density map annotations and test on the target domain on base architecture. (ii) Step learning on target: the base model is pre-trained only with density map on target dataset, and three subnetworks (DAN, LCN, HCN) are fine-tuned with pre-trained base model also using the target dataset. (iii) Fine-tune on target: the base model pre-trained on the source domain is used as the initialization of the entire framework and is fine-tuned on the target domain. The transfer results of MCNN \cite{Zhang:2016fr} is illustrated for comparison. It is clear that w/o fine-tune achieves reasonable performance compared with MCNN \cite{Zhang:2016fr}; fine-tuning on the target domain further improves 5.6 MAE and 6.4 MSE. These indicate that our model is flexible and can transfer between datasets with dynamic scenarios.

\section{Conclusions}
\label{sec:concl}
In this paper, we demonstrate the density adaption networks for crowd counting in dynamic scenarios. The framework leverages a density level estimator to adaptively choose between different counter networks that are explicitly trained for different crowd density domains. Experiments on two major crowd counting benchmarks show promising results of the proposed approach.

\ifCLASSOPTIONcaptionsoff
  \newpage
\fi

\balance
\bibliographystyle{IEEEtran}
\bibliography{cc_dan}

\begin{thebibliography}{10}
\providecommand{\url}[1]{#1}
\csname url@samestyle\endcsname
\providecommand{\newblock}{\relax}
\providecommand{\bibinfo}[2]{#2}
\providecommand{\BIBentrySTDinterwordspacing}{\spaceskip=0pt\relax}
\providecommand{\BIBentryALTinterwordstretchfactor}{4}
\providecommand{\BIBentryALTinterwordspacing}{\spaceskip=\fontdimen2\font plus
\BIBentryALTinterwordstretchfactor\fontdimen3\font minus
  \fontdimen4\font\relax}
\providecommand{\BIBforeignlanguage}[2]{{%
\expandafter\ifx\csname l@#1\endcsname\relax
\typeout{** WARNING: IEEEtran.bst: No hyphenation pattern has been}%
\typeout{** loaded for the language `#1'. Using the pattern for}%
\typeout{** the default language instead.}%
\else
\language=\csname l@#1\endcsname
\fi
#2}}
\providecommand{\BIBdecl}{\relax}
\BIBdecl

\bibitem{2016spl_kzhang}
K.~Zhang, Z.~Zhang, Z.~Li, and Y.~Qiao, ``Joint face detection and alignment
  using multitask cascaded convolutional networks,'' \emph{IEEE Signal
  Processing Letters}, vol.~23, no.~10, pp. 1499--1503, 2016.

\bibitem{wang2017evolving}
L.~Wang, Y.~Lu, H.~Wang, Y.~Zheng, H.~Ye, and X.~Xue, ``Evolving boxes for fast
  vehicle detection,'' in \emph{ICME}, 2017, pp. 1135--1140.

\bibitem{lyu2017ua}
S.~Lyu \emph{et~al.}, ``Ua-detrac 2017: Report of avss2017 \& iwt4s challenge
  on advanced traffic monitoring,'' in \emph{AVSS}, 2017, pp. 1--7.

\bibitem{Zhang:2015id}
C.~Zhang, H.~Li, X.~Wang, and X.~Yang, ``{Cross-scene crowd counting via deep
  convolutional neural networks},'' in \emph{CVPR}, 2015, pp. 833--841.

\bibitem{Zhang:2016fr}
Y.~Zhang, D.~Zhou, S.~Chen, S.~Gao, and Y.~Ma, ``{Single-Image Crowd Counting
  via Multi-Column Convolutional Neural Network},'' in \emph{CVPR}, 2016, pp.
  589--597.

\bibitem{sam2017switching}
D.~B. Sam, S.~Surya, and R.~V. Babu, ``Switching convolutional neural network
  for crowd counting,'' in \emph{CVPR}, vol.~1, no.~3, 2017, p.~6.

\bibitem{Sindagi:2017vv}
V.~A. Sindagi and V.~M. Patel, ``{Generating High-Quality Crowd Density Maps
  using Contextual Pyramid CNNs },'' in \emph{CVPR}, 2017, pp. 1--14.

\bibitem{zhang2017fcn}
S.~Zhang, G.~Wu, J.~P. Costeira, and J.~M. Moura, ``Fcn-rlstm: Deep
  spatio-temporal neural networks for vehicle counting in city cameras,'' in
  \emph{ICCV}, 2017.

\bibitem{Xiong:2017ug}
F.~Xiong, X.~Shi, and D.-Y. Yeung, ``{Spatiotemporal Modeling for Crowd
  Counting in Videos},'' \emph{ICCV}, 2017.

\bibitem{kumagai2017mixture}
S.~Kumagai, K.~Hotta, and T.~Kurita, ``Mixture of counting cnns: Adaptive
  integration of cnns specialized to specific appearance for crowd counting,''
  \emph{arXiv preprint arXiv:1703.09393}, 2017.

\bibitem{arteta2016counting}
C.~Arteta, V.~Lempitsky, and A.~Zisserman, ``Counting in the wild,'' in
  \emph{ECCV}, 2016, pp. 483--498.

\bibitem{Xie:2015wr}
W.~Xie, J.~A. Noble, and A.~Zisserman, ``{Microscopy Cell Counting with Fully
  Convolutional Regression Networks},'' in \emph{MICCAI}, 2015, pp. 1--8.

\bibitem{idrees2013multi}
H.~Idrees, I.~Saleemi, C.~Seibert, and M.~Shah, ``Multi-source multi-scale
  counting in extremely dense crowd images,'' in \emph{CVPR}, 2013, pp.
  2547--2554.

\bibitem{Shang:2016iq}
C.~Shang, H.~Ai, and B.~Bai, ``{End-to-end crowd counting via joint learning
  local and global count},'' in \emph{ICIP}, 2016, pp. 1215--1219.

\bibitem{loy2013crowd}
C.~C. Loy, K.~Chen, S.~Gong, and T.~Xiang, ``Crowd counting and profiling:
  Methodology and evaluation,'' in \emph{Modeling, Simulation and Visual
  Analysis of Crowds}.\hskip 1em plus 0.5em minus 0.4em\relax Springer, 2013,
  pp. 347--382.

\bibitem{sindagi2017survey}
V.~A. Sindagi and V.~M. Patel, ``A survey of recent advances in cnn-based
  single image crowd counting and density estimation,'' \emph{Pattern
  Recognition Letters}, 2017.

\bibitem{li2008estimating}
M.~Li, Z.~Zhang, K.~Huang, and T.~Tan, ``Estimating the number of people in
  crowded scenes by mid based foreground segmentation and head-shoulder
  detection,'' in \emph{ICPR}, 2008, pp. 1--4.

\bibitem{leibe2005pedestrian}
B.~Leibe, E.~Seemann, and B.~Schiele, ``Pedestrian detection in crowded
  scenes,'' in \emph{CVPR}, vol.~1, 2005, pp. 878--885.

\bibitem{wang2009crowd}
L.~Wang and N.~H. Yung, ``Crowd counting and segmentation in visual
  surveillance,'' in \emph{ICIP}, 2009, pp. 2573--2576.

\bibitem{chan2009bayesian}
A.~B. Chan and N.~Vasconcelos, ``Bayesian poisson regression for crowd
  counting,'' in \emph{ICCV}, 2009, pp. 545--551.

\bibitem{chen2012feature}
K.~Chen, C.~C. Loy, S.~Gong, and T.~Xiang, ``Feature mining for localised crowd
  counting,'' in \emph{BMVC}, 2012.

\bibitem{Segui:2015ho}
S.~Segu{\'\i}, O.~Pujol, and J.~Vitri{\`a}, ``{Learning to count with deep
  object features},'' in \emph{CVPR Workshops}, 2015, pp. 90--96.

\bibitem{CarlosArteta:2014vm}
C.~Arteta, V.~Lempitsky, J.~A. Noble, and A.~Zisserman, ``{Interactive Object
  Counting},'' in \emph{ECCV}, 2014, pp. 1--15.

\bibitem{wang2017deep}
Q.~Wang, J.~Wan, and Y.~Yuan, ``Deep metric learning for crowdedness
  regression,'' \emph{IEEE Transactions on Circuits and Systems for Video
  Technology}, 2017.

\bibitem{chan2008privacy}
A.~B. Chan, Z.-S.~J. Liang, and N.~Vasconcelos, ``Privacy preserving crowd
  monitoring: Counting people without people models or tracking,'' in
  \emph{CVPR}, 2008, pp. 1--7.

\bibitem{ryan2009crowd}
D.~Ryan, S.~Denman, C.~Fookes, and S.~Sridharan, ``Crowd counting using
  multiple local features,'' in \emph{Digital Image Computing: Techniques and
  Applications}, 2009, pp. 81--88.

\bibitem{chan2012counting}
A.~B. Chan and N.~Vasconcelos, ``Counting people with low-level features and
  bayesian regression,'' \emph{IEEE Transactions on Image Processing}, vol.~21,
  no.~4, pp. 2160--2177, 2012.

\bibitem{fu2012real}
H.~Fu, H.~Ma, and H.~Xiao, ``Real-time accurate crowd counting based on rgb-d
  information,'' in \emph{ICIP}, 2012, pp. 2685--2688.

\bibitem{lempitsky2010learning}
V.~Lempitsky and A.~Zisserman, ``Learning to count objects in images,'' in
  \emph{NIPS}, 2010, pp. 1324--1332.

\bibitem{walach2016learning}
E.~Walach and L.~Wolf, ``Learning to count with cnn boosting,'' in
  \emph{ECCV}.\hskip 1em plus 0.5em minus 0.4em\relax Springer, 2016, pp.
  660--676.

\bibitem{onoro2016towards}
D.~Onoro-Rubio and R.~J. L{\'o}pez-Sastre, ``Towards perspective-free object
  counting with deep learning,'' in \emph{ECCV}, 2016, pp. 615--629.

\bibitem{yu2017dilated}
F.~Yu, V.~Koltun, and T.~Funkhouser, ``Dilated residual networks,'' \emph{arXiv
  preprint arXiv:1705.09914}, 2017.

\bibitem{zheng2018learning}
Y.~Zheng, H.~Ye, L.~Wang, and J.~Pu, ``Learning multiviewpoint context-aware
  representation for rgb-d scene classification,'' \emph{IEEE Signal Processing
  Letters}, vol.~25, no.~1, pp. 30--34, 2018.

\bibitem{rene2017temporal}
C.~Lea, M.~D. Flynn, R.~Vidal, A.~Reiter, and G.~D. Hager, ``Temporal
  convolutional networks for action segmentation and detection,'' in
  \emph{CVPR}, 2017, pp. 1003--1012.

\bibitem{xu2018dense}
B.~Xu, H.~Ye, Y.~Zheng, H.~Wang, T.~Luwang, and Y.-G. Jiang, ``Dense dilated
  network for few shot action recognition,'' in \emph{ICMR}, 2018, pp.
  379--387.

\end{thebibliography}

\end{document}